\documentclass[runningheads]{llncs}
\usepackage{graphicx}
\usepackage{tikz}
\usepackage{comment}
\usepackage{amsmath,amssymb} % define this before the line numbering.
\usepackage{color}
\usepackage{booktabs}
\usepackage{multirow}
\usepackage{float}
\usepackage[pagebackref,breaklinks,colorlinks]{hyperref}
\usepackage{cases}
\usepackage{mathrsfs}
\usepackage{indentfirst}

\usepackage[accsupp]{axessibility}  % Improves PDF readability for those with disabilities.

\begin{document}
\newcommand{\jiawei}[1]{\textcolor{blue}{Jiawei:#1}}
\newcommand{\fan}[1]{\textcolor{cyan}{\footnote{\scriptsize Fan:#1}}}
% \linenumbers
\pagestyle{headings}
\mainmatter
\def\ECCVSubNumber{4005}  % Insert your submission number here

\title{Densely Constrained Depth Estimator for Monocular 3D Object Detection} % Replace with your title
% \title{Weighting by Matching: Every Keypoint Differs in Monocular 3D Object Detection}
%\begin{comment}
% \titlerunning{ECCV-22 submission ID \ECCVSubNumber} 
% \authorrunning{ECCV-22 submission ID \ECCVSubNumber} 
% \author{Anonymous ECCV submission}
% \institute{Paper ID \ECCVSubNumber}
%\end{comment}
%******************

% % CAMERA READY SUBMISSION

\titlerunning{Densely Constrained Depth Estimator for Monocular 3D Object Detection}
\author{Yingyan Li\inst{1,2,4,5} \and
Yuntao Chen \and
Jiawei He\inst{1,2,4} \and
Zhaoxiang Zhang\inst{1,2,3,4,5}
}
\institute{Institute of Automation, Chinese Academy of Sciences (CASIA) \and
University of Chinese Academy of Sciences (UCAS) \and
Centre for Artificial Intelligence and Robotics, HKISI\_CAS \and
National Laboratory of Pattern Recognition (NLPR) \and
School of Future Technology, UCAS
\email{\tt\small\{liyingyan2021,hejiawei2019,zhaoxiang.zhang\}@ia.ac.cn,\\
chenyuntao08@gmail.com}
}
\authorrunning{Y. Li et al.}


%******************
\maketitle

\begin{abstract}
Estimating accurate 3D locations of objects from monocular images is a challenging problem because of lacking depth. Previous work shows that utilizing the object's keypoint projection constraints to estimate multiple depth candidates boosts the detection performance. However, the existing methods can only utilize vertical edges as projection constraints for depth estimation. So these methods only use a small number of projection constraints and produce insufficient depth candidates, leading to inaccurate depth estimation. In this paper, we propose a method that utilizes \emph{dense} projection constraints from edges of any direction. In this way, we employ much more projection constraints and produce considerable depth candidates. Besides, we present a graph matching weighting module to merge the depth candidates. The proposed method \emph{DCD} (Densely Constrained Detector) achieves state-of-the-art performance on the KITTI and WOD benchmarks. Code is released at \url{https://github.com/BraveGroup/DCD}.
\end{abstract}

\section{Introduction}
\indent Monocular 3D detection~\cite{li2021monocular_km3d,yan2017mono3d,chen2020monopair,DBLP:conf/cvpr/ZhangL021_monoflex} has become popular because images are large in number, easy to obtain, and have dense information. Nevertheless, the lack of depth information in monocular images is a fatal problem for 3D detection. Some methods~\cite{brazil2019m3d_rpn,liu2020smoke} use deep neural networks to regress the 3D bounding boxes directly, but it is challenging to estimate the 3D locations of the objects from 2D images. Another line of work~\cite{weng2019monocular_pseudo_lidar,ding2020learning_d4lcn,dd3d} employs a pre-trained depth estimator. However, training the depth estimator is separated from the detection part, requiring a large amount of additional data. In addition, some works~\cite{li2021monocular_km3d,DBLP:conf/cvpr/ZhangL021_monoflex,liu2021autoshape} use geometric constraints, i.e., regresses the 2D/3D edges, and then estimates the object's depth from the 2D-3D edge projection constraints. These works employ 3D shape prior information and exhibit state-of-the-art performance, which is worthy of future research.

\begin{figure*}
\begin{center}
\includegraphics[width=0.85\linewidth]{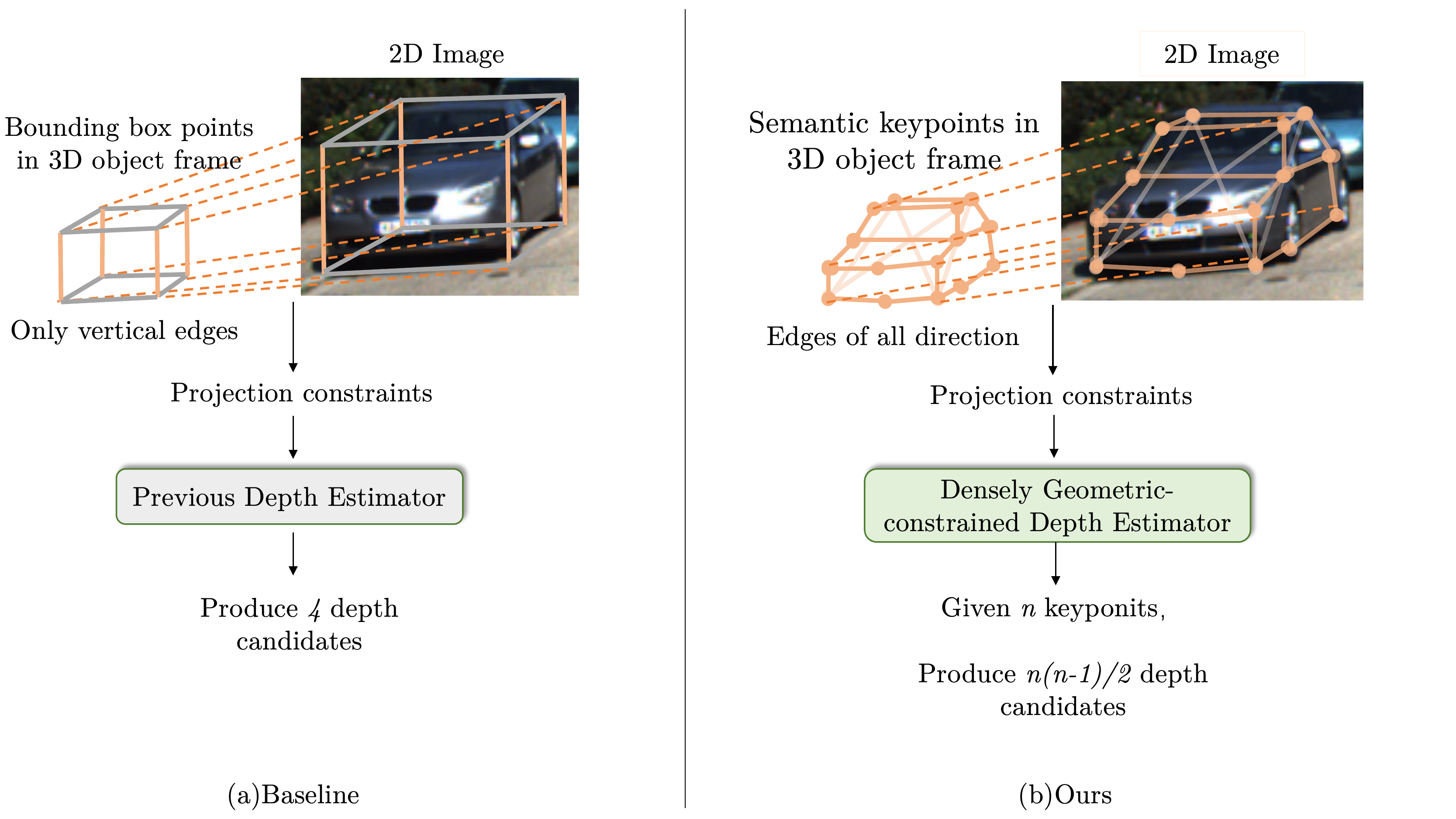}
\caption{The object's depth is estimated by 2D-3D edge projection constraints. This figure compares the involved edges in the object's depth estimation between (a) previous work and (b) ours. The previous work only deals with vertical edges. Our work is able to handle the edges of any direction.}
\label{fig:fig1}
\end{center}
\end{figure*}

A problem of the previous work is that their geometric constraints are insufficient. Specifically, some existing methods~\cite{DBLP:conf/cvpr/ZhangL021_monoflex,lu2021gupnet,zhang2021project_depth} estimate the height of the 2D bounding box and the 3D bounding box, and then generate the depth candidates of an object from 2D-3D height projection constraints. The final depth is produced by weighting all the depth candidates. As Fig.~\ref{fig:fig1} shows, this formulation is only suitable for the vertical edges, which means they only utilize a tiny amount of constraints and 3D prior, leading to inaccurate depth estimations.\par

Some of the depth candidates are of low quality, so weighting is needed. However, the previous work's weighting methods are suboptimal. Since the final depth is derived from the weighted average of depth candidates, the weight should reflect the quality of each depth candidate. Existing methods~\cite{liu2021autoshape} use a branch to regress the weight of each depth candidate directly, and this branch is paralleled to the keypoints regression branch. So the weighting branch does not know each keypoint's quality. Some work predicts the uncertainty of each depth to measure the quality of the depth and use the uncertainty to weight~\cite{lu2021gupnet,DBLP:conf/cvpr/ZhangL021_monoflex}. However, they obtain the uncertainty of each depth candidate independently, and they do not supervise the weight explicitly.\par

To address the problem of insufficient geometric constraints, we propose a \emph{Densely Geometric-constrained Depth Estimator (DGDE)}. \emph{DGDE} can estimate depth candidates from projection constraints provided by edges of any direction, no more limited to the vertical edges. This estimator allows better use of the shape information of the object. In addition, training the neural network with abundant  2D-3D projection constraints helps the neural network understand the mapping relationship from the 2D plane to the 3D space.\par

To weight the depth candidates properly, we propose a new depth candidates weighting module that employs graph matching, named \emph{Graph Matching Weighting module}.
We construct \emph{complete graphs} based on 2D and 3D semantic keypoints. In a 2D keypoint graph, the 2D keypoint coordinates are placed on the vertices, and an edge represents a pair of 2D keypoints. The 3D keypoint graph is constructed in the same way. We then match the 2D edges and 3D edges and produce the matching scores. The 2D-3D edge matching score is used as the weight of the corresponding depth candidate. These weights are explicitly supervisable. Moreover, the information of the entire 2d/3d edges is used to generate each 2d-3d edge matching score.

In summary, our main contributions are: 
\begin{enumerate}
\item We propose a \emph{Dense Geometric-constrained Depth Estimator (DGDE)}. Different from the previous methods, \emph{DGDE} estimates depth candidates utilizing projection constraints of edges of any direction. Therefore, considerable 2D-3D projection constraints are used, producing considerable depth candidates. We produce high-quality final depth based on these candidates.

\item We propose an effective and interpretable \emph{Graph Matching Weighting module (GMW)}. We construct the 2D/3D graph from 2D/3D keypoints respectively. Then we regard the graph matching score of the 2D-3D edge as the weight of the corresponding depth candidate. This strategy utilizes all the keypoints' information and produces explicitly supervised weights.

\item We localize each object more accurately by weighting the estimated depth candidates with corresponding matching scores. Our \emph{Densely Constrained Detector (DCD)} achieves state-of-the-art performance on the KITTI and Waymo Open Dataset (WOD) benchmarks.
\end{enumerate}

\section{Related Work}
\noindent{\bf Monocular 3D Object Detection.}
Monocular 3D object detection ~\cite{kundu20183d_3drcnn,chabot2017deepmanta,chen2020monopair,grabner20183d_shapeprior} becomes more and more popularity because monocular images are easy to obtain. The existing methods can be divided into two categories: single-center-point-based and multi-keypoints-based. \\
\indent Single-center-point-based methods ~\cite{liu2020smoke,zhou2019objects_centernet,zhou2020iafa} use the object's center point to represent an object. In detail, M3D-RPN~\cite{brazil2019m3d_rpn} proposes a depth-aware convolutional layer with the estimated depth. MonoPair~\cite{chen2020monopair} discovers that the relationships between nearby objects are useful for optimizing the final results. Although single-center-point-based methods are simple and fast, location regression is unstable because only one center point is utilized. Therefore, the multi-keypoints-based methods~\cite{liu2021autoshape,DBLP:conf/cvpr/ZhangL021_monoflex,li2021monocular_km3d} are drawing more and more attention recently.\\
\indent Multi-keypoints-based methods predict multiple keypoints for an object. More keypoints provide more projection constraints. The projection constraints are useful for training the neural network because constraints build the mapping relationship from the 2D image plane to the 3D space. Deep MANTA~\cite{chabot2017deepmanta} defines 4 wireframes as templates for matching cars, while 3D-RCNN ~\cite{kundu20183d_3drcnn} proposes a render-and-compare loss.  Deep3DBox \cite{mousavian20173d_deep3dbox} utilizes 2D bounding boxes as constraints to refine 3D bounding boxes. KM3D \cite{li2021monocular_km3d} localizes the objects utilizing eight bounding boxes points projection. MonoJSG~\cite{lian2022monojsg} constructs an adaptive cost volume with semantic features to model the depth error. AutoShape~\cite{liu2021autoshape}, highly relevant to this paper, regresses 2D and 3D semantic keypoints and weighs them by predicted scores from a parallel branch. MonoDDE~\cite{MonoDCD} is a concurrent work. It uses more geometric constraints than MonoFlex~\cite{DBLP:conf/cvpr/ZhangL021_monoflex}. However, it only uses geometric constraints based on the object's center and bounding box points, while we use dense geometric constraints derived from semantic keypoints. In this paper, we propose a novel edge-based depth estimator. The estimator produces depth candidates by projection constraints, consisting of edges of any direction. We weight each edge by graph matching to derive the final depth of the object.\par

\noindent{\bf Graph Matching and Message Passing.}
Graph matching~\cite{GM_survey} is defined as matching vertices and edges between two graphs. This problem is formulated as a Quadratic Assignment Problem (QAP)~\cite{QAP} originally. It has been widely applied in different tasks, such as multiple object tracking~\cite{He_2021_CVPR}, semantic keypoint matching~\cite{zhou2012factorized} and point cloud registration~\cite{fu2021robust}. In the deep learning era, graph matching has become a differentiable module. The spectral relaxation~\cite{zanfir2018deep}, quadratic relaxation~\cite{He_2021_CVPR} and Lagrange decomposition~\cite{rolinek2020deep} are common in use. However, one simple yet effective way to implement the graph matching layer is using the Sinkhorn~\cite{campbell2020solving,sarlin2020superglue,wang2019learning} algorithm, which is used in this paper. This paper utilizes a graph matching module to 
achieve the message passing between keypoints. This means that we calculate the weighting score not only from the keypoint itself but also by taking other keypoints' regression quality into consideration, which is a kind of message passing between keypoints. All the keypoints on the object help us judge each keypoint's importance from the perspective of the whole object. Message passing~\cite{gilmer2017neural} is a popular design, e.g., in graph neural network~\cite{battaglia2018relational} and transformer~\cite{vaswani2017attention}. The message passing module learns to aggregate features between nodes in the graph and brings the global information to each node. In the object detection task, Relation Networks~\cite{hu2018relation} is a pioneer work utilizing message passing between proposals. Message passing has also been used for 3D vision. Message passing between frames~\cite{yang20213d}, voxels~\cite{fan2021embracing} and points~\cite{sheng2021improving} is designed for LiDAR-based 3D object detection. However, for monocular 3D object detection, message passing is seldom considered. In recent work, PGD~\cite{wang2022probabilistic} conducts message passing in the geometric relation graph. But it does not consider the message passing within the object.

\begin{figure*}
\centering
\includegraphics[width=\linewidth]{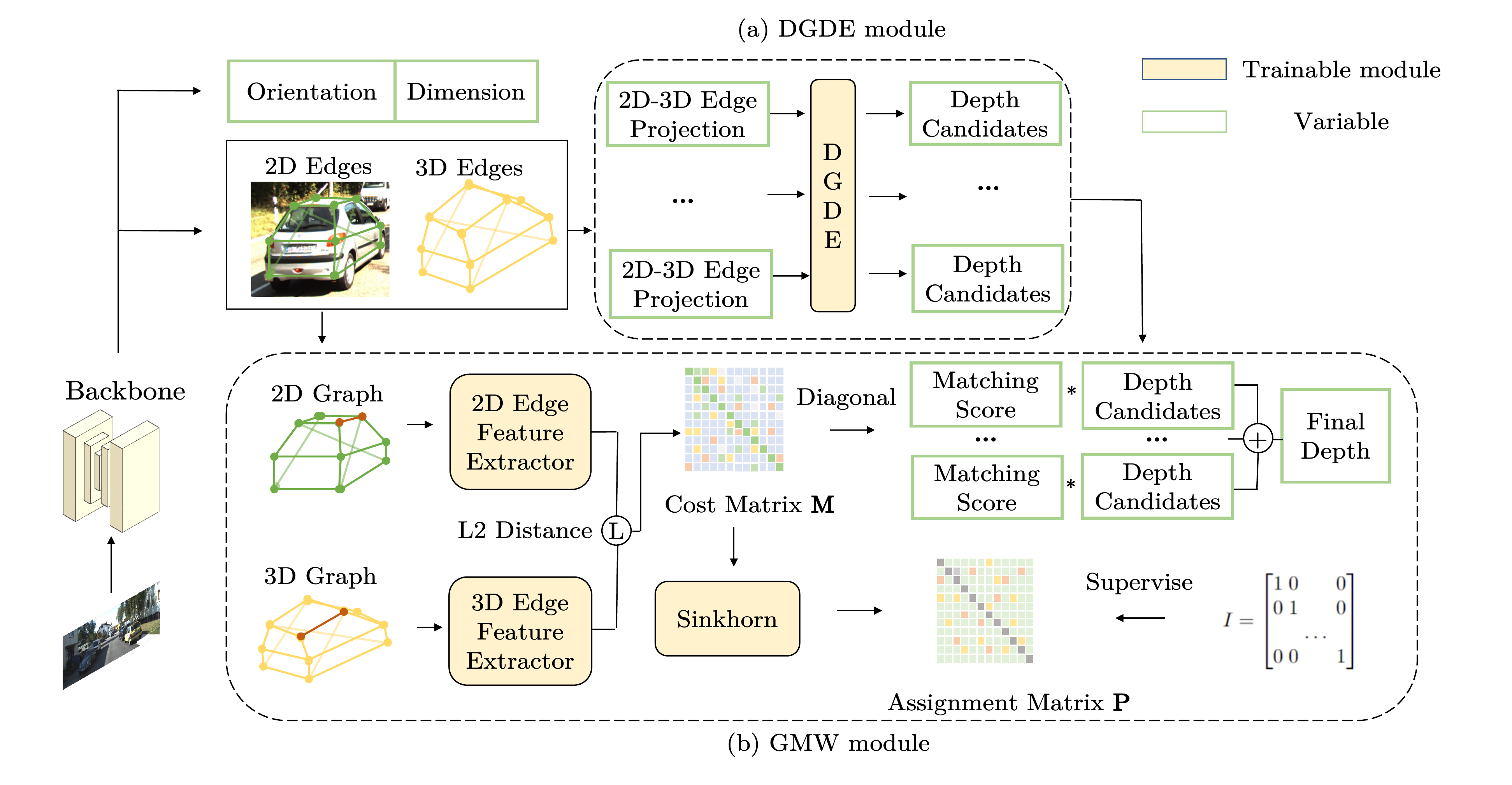}
\caption{Overview of our framework. (a) We propose a method \emph{Densely Geometric-constrained Depth Estimator} (DGDE). DGDE is able to estimate the object's depth candidates from 2D-3D projection constraints of edges of any direction. (b) \emph{Graph Matching Weighting module} (GMW) obtains the weights of estimated depth candidates by Graph Matching. A robust depth is derived from combing the multiply depth candidates with corresponding weights.}
\label{fig:framework}
\end{figure*}

\section{Methodology}
The overview of our framework is in Fig.~\ref{fig:framework}. We employ a single-stage detector \cite{DBLP:conf/cvpr/ZhangL021_monoflex} to detect the object's 3D attributes from the monocular image. We propose the \emph{Densely Geometric-constrained Depth Estimator (DGDE)}, which can calculate the depth from any direction' 2D-3D edge. The \emph{DGDE} can effectively utilize the semantic keypoints of the object and produce many depth candidates. Besides, we utilize the regressed 2D edges, 3D edges, and orientation as the input for our 2D-3D edge Graph Matching network. Our \emph{Graph Matching Weighting module (GMW)} matches each 2D-3D edge and produces a matching score. By combining the multiple depths with their corresponding matching scores, we can finally generate a robust depth for the object. 

\subsection{Geometric-based 3D Detection Definition}\label{sec:definition}
The geometric-based monocular 3D object detection estimates the object's location by 2D-3D projection constraints. Specifically, the network predicts the object's dimension $(h,w,l)$, rotation $r_y$, since autonomous driving datasets generally assume that the ground is flat. Assuming an object has $n$ semantic keypoints, we regress the $i$-th$ (i=1,2,\dots,n)$ keypoint's 2D coordinate $(u^i,v^i)$ in image coordinate and 3D coordinate $(x_o^i,y_o^i,z_o^i)$  in object frame. The object frame's coordinate origin is the object's center point. The $i$-th 2D-3D keypoint projection constraint is established from $(u^i,v^i,x_o^i,y_o^i,z_o^i,r_y)$. Given $n$ semantic 2D-3D keypoint projection constraints, it is an overdetermined problem for solving 3D object location $(x_c,y_c,z_c)$, which is the translation vector for transforming the points from the object frame into the camera frame. The method of generating semantic keypoints of each object is adapted from \cite{liu2021autoshape}. We establish a few car models by PCA and refine the models by the 3D points segmented from the points cloud and 2D masks. After we obtain the keypoints, we can use our \emph{DGDE} to estimate the object's depth from the keypoint projection constraints.

\subsection{Densely Geometric-constrained Depth Estimation}
While previous depth estimation methods~\cite{DBLP:conf/cvpr/ZhangL021_monoflex} only take vertical edges into account, our \emph{DGDE} can handle edges of any direction. Therefore, we are able to utilize much more constraints to estimate the depth of each depth candidate.

Next, we will show the details of estimating dense depth candidates of an object from 2D-3D keypoint projection constraints. The solution is based on the keypoint's projection relationship from 3D space to the 2D image. The $i$-th$(i=1,2,\dots,n)$ keypoint's 3D coordinate $(x_o^i,y_o^i,z_o^i)$ is defined in the object frame and is projected on a 2D image plane by the equation:
\begin{equation}\label{projection_eq}
s_i[u_i,v_i,1]^T=\textbf{K}[\textbf{R} |\textbf{t} ][x^i_o,y^i_o,z^i_o,1]^T,
\end{equation}
where $s_i$ is the $i$-th keypoint's depth, $\textbf{K}$ is the camera intrinsic matrix and $\textbf{K},\textbf{R},\textbf{t}$ is represented as:
\begin{equation}\label{camera_matrix}
\textbf{K}=\begin{bmatrix}
f_x & 0   & c_x\\
0   & f_y & c_y\\
0   & 0   & 1
\end{bmatrix},
\textbf{R}= \begin{bmatrix}
\cos{r_y} & 0   & \sin{r_y}\\
0   & 1 & 0\\
-\sin{r_y}   & 0   & \cos{r_y}
\end{bmatrix},
\textbf{t}=[x_c,y_c,z_c]^t.
\end{equation}
By Eq.~\eqref{projection_eq} and Eq.~\eqref{camera_matrix}, the equation of $i$-th keypoint's projection constraint is denoted as:
\begin{equation}\label{eq:uxyz_i}
\begin{cases}
s_{i}=z_c-x^i_o\sin{r_y}+z^i_o\cos{r_y}, \\
\tilde{u}_i(z_c-x^i_o\sin{r_y}+z^i_o\cos{r_y}) = x_c+x^i_o\cos{r_y}+z^i_o\sin{r_y}, \\ 
\tilde{v}_i(z_c-x^i_o\sin{r_y}+z^i_o\cos{r_y}) = 
y_c+y^i_o,\\   
\end{cases}
\end{equation}
where $\tilde{u_i}=\frac{u_i-c_x}{f_x},\tilde{v_i}=\frac{v_i-c_y}{f_y} $. Intuitively, $(z_c-x^i_o\sin{r_y}+z^i_o\cos{r_y})$ means an object's $i$-th 3D keypoint's  $Z$ coordinate (i.e., depth) in the camera coordinate.  $(x_c+x^i_o\cos{r_y}+z^i_o\sin{r_y})$ means 3D keypoint's X coordinate while $(y_c+y^i_o)$ means its Y coordinate.

Similarly, the $j$-th$(j=1,2,\dots,n)$ projection constraint is denoted as:
\begin{equation}\label{eq:uxyz_j}
\begin{cases}
s_{j}=z_c-x^j_o\sin{r_y}+z^j_o\cos{r_y}, \\
\tilde{u}_j(z_c-x^j_o\sin{r_y}+z^j_o\cos{r_y}) = x_c+x^j_o\cos{r_y}+z^j_o\sin{r_y}, \\ 
\tilde{v}_j(z_c-x^j_o\sin{r_y}+z^j_o\cos{r_y}) = 
y_c+y^j_o.\\ 
\end{cases}
\end{equation}

From Eq.~\eqref{eq:uxyz_i} and Eq.~\eqref{eq:uxyz_j}, we can densely obtain the $z_c$ from the $i$-th,$j$-th,$i\neq j$ keypoint(i.e., $edge_{ij}$) projection constraints as:
\begin{numcases}{z_c^{ij}=}
    \frac{l_i-l_j}{\tilde{u_i}-\tilde{u_j}},  \\
    \frac{h_i-h_j}{\tilde{v_i}-\tilde{v_j}},\label{eq:zij}
\end{numcases}
where 
$l_i=x_o^i\cos{(r_y)}+z_o^i\sin{(r_y)}+u^i(x_o^i\sin{(r_y)}-z_o^i\cos{(r_y)})$ and $h_i=y_o^i+v^i(x_o^i\sin{(r_y)}-z_o^i\cos{(r_y)}).$
This equation reveals that depth can be calculated by the projection constraints of an edge of any direction. Given $z_c$, we can estimate $x_c,y_c$ from Eq.~\eqref{eq:uxyz_i} as $x_c^i=u_i z_c-l_i,y_c^i=v_i z_c-h_i$.

We generate $m=n(n-1)/2$ depth candidates given $n$ keypoints.  It is inevitable to meet some low-quality depth candidates in such a large number of depths. Therefore, an appropriate weighting method is necessary to ensemble these depth candidates.

\subsection{Depth Weighting by Graph Matching}
As we estimate the depth candidate $z_c^{ij} (i,j=1,\cdots,n)$ for the object $o$ from \emph{DGDE}, the final depth $z_c$ of the object can be weighted from these depth estimations according to the estimation quality $w_{i,j}$, as
\begin{equation}
    z_c = \sum_{i < j} w_{i,j}  z_c^{ij}.
    \label{eq:weightedsum}
\end{equation}
In this section, we propose a new weighting method, called \emph{Graph Matching Weighting module (GMW)}.\par
\noindent{\bf Graph Construction and Edge Feature extraction.} We construct 2D keypoint graph $\mathcal{G}_{2d}=(\mathcal{V}_{2d}, \mathcal{E}_{2d})$ and 3D keypoint graph $\mathcal{G}_{3d}=(\mathcal{V}_{3d}, \mathcal{E}_{3d})$. In $\mathcal{G}_{2d}$, each vertex $i\in\mathcal{V}_{2d}$ denotes a predicted keypoint  $^{(2d)}\mathbf{p}^i=[u^i,v^i]$ in image coordinate and the edge $^{(2d)}e^{i,j}\in \mathcal{E}_{2d}$ denotes the pair of $^{(2d)}\mathbf{p}^i$ and $^{(2d)}\mathbf{p}^j$, the edge feature $^{(2d)}\mathbf{f}^{i,j}$ is extracted from the 2D coordinate $^{(2d)}\mathbf{p}^i$ and $^{(2d)}\mathbf{p}^j$. The 3D keypoint graph is almost similar to the 2D keypoint graph. The only difference is that the vertex $i'\in\mathcal{V}_{3d}$ denotes the 3D coordinate $^{(3d)}\mathbf{p}^{i'}=[x_o^{i'},y_o^{i'},z_o^{i'}]$. 

Following~\cite{yi2018learning_to_find}, the 2D and 3D edge feature extractor are as
\begin{align}
    ^{(2d)}\mathbf{f}^{i,j}_{k}&=\mathtt{ReLU}_k(\mathtt{BN}_k(\mathtt{CN}_k(^{(2d)}\mathtt{FC}_k(^{(2d)}\mathbf{f}^{i,j}_{k-1})))),\\
    ^{(3d)}\mathbf{f}^{i',j'}_{k}&=\mathtt{ReLU}_k(\mathtt{BN}_k(\mathtt{CN}_k(^{(3d)}\mathtt{FC}_k(^{(3d)}\mathbf{f}^{i',j'}_{k-1})))),
\end{align}  
where $k\in\{1,\cdots,K\}$ denotes the index of layers, and $\mathtt{FC},\mathtt{CN},\mathtt{BN},\mathtt{ReLU}$ denote fully-connected layer, Context
Normalization~\cite{yi2018learning_to_find}, Batch Normalization, and ReLU, respectively. It is worth mentioning that Context
Normalization extracts the global information of all edges. The input of the edge feature extractor is $\mathbf{f}^{i,j}_0=[\mathbf{p}^i,\mathbf{p}^j]$, where $[\cdot]$ denotes the concatenation of the vectors. 
The output of edge feature extractor $^{(2d)}\mathbf{f}^{i,j}$ and $^{(3d)}\mathbf{f}^{i',j'}$ should be L2-normalized to $[0,1]$.

\noindent{\bf Graph matching layer.}
Given the extracted 2D and 3D edge features, the Cost Matrix $\mathbf{M}\in \mathbb{R}^{m\times m}$ is calculated from the L2 distance between each 2D edge feature $^{(2d)}\mathbf{f}^{i,j}$ on the edge $s$ and 3D edge feature $^{(3d)}\mathbf{f}^{i',j'}$ on the edge $t$:
\begin{equation}
\mathbf{M}_{s,t}= \mathtt{L2}(^{(2d)}\mathbf{f}^{i,j}, ^{(3d)}\mathbf{f}^{i',j'}),
(s,t\in \{1,\cdots,m\}),
\end{equation}
where $m$ denotes the number of edges. Then we take $\textbf{M}$ as the input of declarative Sinkhorn layer\cite{campbell2020solving} to gain the Assignment Matrix $\textbf{P}$. The Sinkhorn layer iteratively optimizes $\textbf{P}$ by minimizing the objective function:
\begin{equation}
\mathcal{F}(\textbf{M},\textbf{P})=\sum\limits_{s=1}^{m}\sum\limits_{t=1}^{m}(\textbf{M}_{s,t}\textbf{P}_{s,t}+ \alpha \textbf{P}_{s,t}(\log \textbf{P}_{s,t}-1)), 
\end{equation}
where $\textbf{P}_{s,t}(\log \textbf{P}_{s,t}-1)$ is a regularization term and $\alpha$ is the coefficient.
$\textbf{P}\in U(\textbf{a},\textbf{b})$ as:
\begin{equation}
U(\textbf{a},\textbf{b})=\{\textbf{X} \in \mathbb{R}^{m\times m}_{+} \vert \textbf{X}\textbf{1}^m = \textbf{1}^m, \textbf{X}^T \textbf{1}^m = \textbf{1}^m\},
\end{equation}
where $\textbf{1}^m$ is an m-dimensional vector with all values to be 1. Note that, Sinkhorn is a differentiable graph matching solver that can make the whole pipeline learnable. When calculating the final depth $z_c$ according to Eq.~\eqref{eq:weightedsum}, the weight $\textbf{w}=\mathtt{Softmax}({1}/Diag(\mathbf{M}))$, where $Diag(\mathbf{M})$ means the vector consisting of diagonal elements of matrix $\mathbf{M}$. Intuitively, it means that we take the similarity of the 2D and 3D edges with the same semantic label as the prediction quality.

\noindent{\bf Loss function.} We design regression loss $\mathcal{L}_m^r$ to supervise the final weighted depth $z_c$, and classification loss $\mathcal{L}_m^c$ to supervise the assignment matrix $\mathbf{P}$ of graph matching output to be an identity matrix. Specifically, $\mathcal{L}_m^c$ is the Binary Entropy Loss (BCE), $\mathcal{L}_m^r$ is an L1 loss:
\begin{equation}
\mathcal{L}^c_m=\sum\limits_{s=1}^m\sum\limits_{t=1}^m BCE(\textbf{P}_{s,t},~\textbf{P}_{s,t}^*),\mathcal{L}^r_m=  |{\sum_{i<j} w_{i,j} z_c^{ij} - z_c^*}|,
\end{equation}
where $\textbf{P}^*=\mathbf{I}$ is the ground truth assignment matrix, $z_c^*$ is the ground truth depth of the object.
The final matching loss is $\mathcal{L}_m=\mathcal{L}^c_m + \beta\mathcal{L}^r_m$, where $\beta$ is a hyper-parameter.

\section{Experiments}
\subsection{Setup}
\noindent{\textbf{Dataset.}}
We evaluate our method on the KITTI~\cite{geiger2013vision_kitti} and Waymo Open Dataset v1.2 (WOD)~\cite{waymo}. The KITTI~\cite{geiger2013vision_kitti} dataset is collected from Europe Streets. It consists of 7481 images for training and 7518 images for testing. We divide the training data into a train set (3712 images) and a validation set (3769 images) as in \cite{zhou2018voxelnet}. Waymo Open Dataset (WOD)~\cite{waymo} has 798 training sequences, 202 validation sequences, and 150 test sequences. The dataset contains images captured by 5 high-resolution cameras in complex environments and is much more challenging than KITTI~\cite{geiger2013vision_kitti} dataset. We only use the images from the FRONT camera for training and evaluation. The training set has 158,081 images and the validation set has 39,848 images.

\noindent{\textbf{Evaluation Metrics.}}
For the KITTI dataset, we compare our methods with previous methods on the \emph{test} set using $AP_{3D|R_{40}}$ result from the test server. In ablation studies, the results on \emph{val} set are reported.
For the WOD, we focus on the category of vehicle. Following the official evaluation criteria, we compare the performance with the state-of-the-art methods using average precision (AP) and average precision weighted by heading (APH) metrics. The results are shown in Table~\ref{tab:my-table}. The objects are classified into two difficulty levels (LEVEL\_1, LEVEL\_2) according to the object's points' number under the LiDAR sensor. 

\subsection{Implementation Details}
\noindent{\textbf{Detection Framework.}}
We apply the 3D object detection framework following \cite{DBLP:conf/cvpr/ZhangL021_monoflex}, which uses DLA-34 \cite{yu2018deep_dla} as backbone. We use MultiBin loss \cite{mousavian20173d_deep3dbox} for rotation. L1 loss is adopted to estimate dimension, 2D/3D keypoints and depth.

\noindent{\textbf{Keypoints.}}
The source of keypoints is discussed in Sec.~\ref{sec:definition}. We use 73 keypoints in total consisting of the following parts: (1) 63 semantic keypoints; (2) 8 bounding box corners and the top center and the bottom center of the 3D bounding box. 
There are 2628 unique keypoint pairs that can be generated from 73 keypoints, so we can obtain 2628 depth estimations at most for each object. For robustness, we select 1500 depth estimations as the final candidates for weighting. The details of the selection strategy are in Sec.~\ref{sec:num_pairs}.

\noindent{\textbf{Training and Inference.}}For the KITTI dataset, all the input images are padded into $1280\times384$. We train the model using AdamW \cite{loshchilov2017decoupled_adamw} optimizer with an initial learning rate of 3e-4 for 100 epochs. The learning rate decays by $10\times$ at 80 and 90 epochs. We train the model on 2 RTX2080Ti GPUs and the batch size is 8. We train the weighting network (i.e., matching network) separately. The weighting network employs the AdamW optimizer with learning rate 1e-4 and weight decay 1e-5.  We first train using the classification loss in the weighting network for 50 epochs and add the regression loss for another 50 epochs. During inference, only monocular images are needed.
For the WOD, the input size of the images is 1920$\times$1280. We ignore objects whose 2D bounding box's width or height is less than 20 pixels. We train our detection model for 20 epochs with 8 RTX2080Ti GPUs. The batch size is 8 and the learning rate is set to 8e-5, decayed $10\times$ at the $18$-th epoch. The rest of the experiment settings are the same as KITTI.

\begin{table*}[t]
\begin{center}
\caption{The result on the KITTI test server compared with other public methods in recent years.}
\label{Test_result}
\resizebox{0.9\textwidth}{!}{
    \begin{tabular}{l|c|c|ccc|ccc}
    \toprule
    \multirow{2}{*}{Methods} &\multirow{2}{*}{Reference} & \multirow{2}{*}{Category}  &
    \multicolumn{3}{c|}{\textit{$AP_{3D|R40|IoU@0.7}$}} & \multicolumn{3}{c}{\textit{$AP_{BEV|R40|IoU@0.7}$}} \\
     & &  & Easy & Mod. & Hard  & Easy & Mod. & Hard \\
    \midrule

    PatchNet \cite{ma2020rethinking_patchnet} &ECCV20 & \multirow{3}{*}{Pretrained Depth}   & 15.68 & 11.12 & 10.17& 22.97 & 16.86 & 14.97 \\

    D4LCN  \cite{ding2020learning_d4lcn} &CVPR20&    & 16.65 & 11.72 & 9.51& 22.51 & 16.02 & 12.55\\
    
    DDMP-3D  \cite{ddmp3d}& CVPR21 &    & 19.71 & 12.78 & 9.80& 28.08 & 17.89 & 13.44\\
    
    \midrule

    CaDDN  \cite{reading2021categorical_caddn} &CVPR21&\multirow{1}{*}{LiDAR Auxiliary}  & 19.17 & 13.41 & 11.46& 27.94 & 18.91 & 17.19\\

    \midrule
    
    RTM3D  \cite{li2020rtm3d}&ECCV20 & \multirow{6}{*}{Directly Regress}   & 14.41 & 10.34 & 8.77& 19.17 & 14.20 & 11.99\\
    
    Movi3D  \cite{movi3d}&ECCV20 &     & 15.19 & 10.90 & 9.26& 22.76 & 17.03 & 14.85\\

    Ground-Aware  \cite{liu2021ground_aware}&RAL21 &     & 21.65 & 13.25 & 9.91& 29.81 & 17.98 & 13.08 \\

    MonoDLE  \cite{monodle}&CVPR21 &     & 17.23 & 12.26 & 10.29& 24.79 & 18.89 & 16.00\\

    MonoRCNN  \cite{monorcnn} &ICCV21&     & 18.36 & 12.65 & 10.03& 25.48 & 18.11 & 14.10 \\
    
    MonoEF \cite{zhou2021monoef}& CVPR21 &     & 21.29 & 13.87 & 11.71& 29.03 & 19.70 & 17.26\\

    \midrule
% \multirow{2}{*}{}
    MonoRUn  \cite{chen2021monorun}&CVPR21 &\multirow{5}{*}{Geometric-based}   & 19.65 & 12.30 & 10.58& 27.94 & 17.34 & 15.24\\
    AutoShape \cite{liu2021autoshape} &ICCV21&   & 22.47 & 14.17 & 11.36& 30.66 & 20.08 & 15.59\\
    GUPNet  \cite{lu2021gupnet} &ICCV21&   & 22.20 & 15.02 & 13.12& 30.29 & 21.19 & 18.20 \\
    MonoFlex(Baseline)  \cite{DBLP:conf/cvpr/ZhangL021_monoflex} &CVPR21&     & 19.94 & 13.89 & 12.07& 28.23 & 19.75 & 16.89\\
    
    \textbf{DCD(Ours)} & ECCV22 & 
    & \textbf{23.81} & \textbf{15.90} & \textbf{13.21}& \textbf{32.55} & \textbf{21.50} & \textbf{18.25}\\
    \bottomrule
    \end{tabular}}
\end{center}
\end{table*}
\begin{figure*}[t]
\begin{center}
\includegraphics[width=\textwidth]{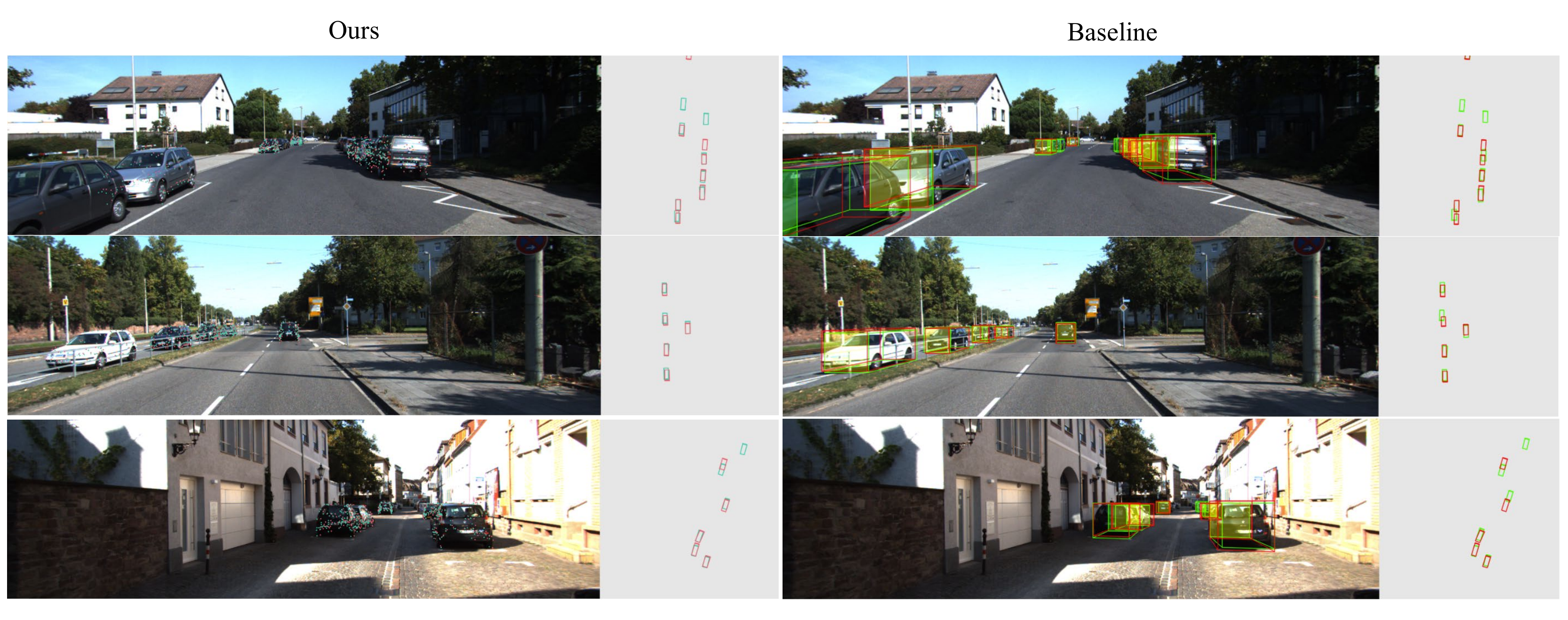}
\caption{The qualitative results on the KITTI \emph{val} set. Rather than representing an object as a bounding box, we utilize semantic keypoints to represent an object. The red boxes represent the ground truth, while the green boxes represent the prediction.}
\label{fig:vis}
\end{center}
\end{figure*}

\subsection{Comparison with State-of-the-art Methods}
\indent Table \ref{Test_result} shows the comparison with other state-of-the-art methods on the KITTI \cite{geiger2013vision_kitti} test set. Our method (\emph{DCD}) achieves state-of-the-art performance on both $AP_{3D}$ and $AP_{BEV}$. We surpass the directly-regress-depth-based method and pretrain-depth-estimator-based methods by a large margin. It reveals that geometric constraints are critical to accurately locating objects. Compared with other geometric-based methods, we still have a significant improvement. We outperform our baseline MonoFlex~\cite{DBLP:conf/cvpr/ZhangL021_monoflex} 
by \textbf{2.01} in $AP_{3D}$ Mod. level, which reveals the importance of sufficient geometric constraints 
for monocular 3D detection. We also surpass other geometric-based methods such as AutoShape~\cite{liu2021autoshape} and GUPNet~\cite{lu2021gupnet} thanks to the dense geometric constraints and effective weighting method. The result of Pedestrian and Cyclist is in Table~\ref{tab:ped_cyc}. Using only bounding box corners as input, we can still observe an improvement over 
MonoFlex~\cite{DBLP:conf/cvpr/ZhangL021_monoflex}. It shows that our method can handle the problem that some objects are hard to obtain semantic keypoints (without CAD models or non-rigid). The qualitative result is in Fig.~\ref{fig:vis}.
% \multicolumn{6}{c}{\textit{Test} set, AP_{3D|R40}}    \\ \cmidrule{2-7}

% \begin{wraptable}[14]{r}{.65\linewidth}
\begin{table}[]
\centering
\caption{The \textit{$AP_{3D|R40}$} results for Pedestrian and Cyclist on KITTI \textit{test} set. We use the bounding box corners as the input of \emph{DGDE}.}
\label{tab:ped_cyc}
\begin{tabular}{c|ccc|ccc}
\toprule
\multirow{2}{*}{Method} 
& \multicolumn{3}{c|}{Pedestrian} & \multicolumn{3}{c}{Cyclist} \\ 
& Easy      & Mod.      & Hard     & Easy     & Mod.     & Hard    \\ \midrule
M3D-RPN~\cite{brazil2019m3d_rpn}                  & 4.92      & 3.48     & 2.94     & 0.94     & 0.65    & 0.47    \\
MonoPair~\cite{chen2020monopair}                   & 10.02      & 6.68     & 5.53     & 3.79     & 2.12    & 1.83    \\
MonoFlex~\cite{DBLP:conf/cvpr/ZhangL021_monoflex}                 & 9.43     & 6.31     & 5.26     & 4.17     & 2.35    & 2.04    \\ 
AutoShape~\cite{liu2021autoshape}                     & 5.46      & 3.74     & 3.03     & \textbf{5.99 }    & \textbf{3.06}    & \textbf{ 2.70}    \\ \midrule
Ours                     & \textbf{10.37}      & \textbf{6.73}     & \textbf{6.28}     & 4.72    & 2.74    & 2.41    \\ \bottomrule
\end{tabular}
\end{table}
% \end{wraptable}

We also achieve state-of-the-art performance on WOD~\cite{waymo} as Table~\ref{tab:my-table} shows. We surpass the previous state-of-the-art methods such as CaDDN~\cite{reading2021categorical_caddn} and PCT~\cite{pct}. We also re-implement MonoFlex~\cite{DBLP:conf/cvpr/ZhangL021_monoflex} on WOD~\cite{waymo} for a fair comparison. Compared with the baseline method, we improve the AP IoU@0.7 by \textbf{0.87}.

\begin{table}[]
\centering
\caption{The result on WOD~\cite{waymo} \emph{val} set. \emph{Italics}: These methods utilize the whole \emph{train} set, while the others use $1/3$ amount of images in \emph{train} set. \ddag: M3D-RPN is re-implemented by \cite{reading2021categorical_caddn}. \dag: PatchNet is re-implemented by \cite{pct}. $^{\ast}$: MonoFlex is our baseline and re-implemented ourselves.}
\label{tab:my-table}
\resizebox{\linewidth}{!}{
\begin{tabular}{c|c|c|c|c|c}
\toprule
  Difficulty &
  Method &
  3D AP (IoU@0.7) &
  3D APH (IoU@0.7) &
  3D AP (IoU@0.5) &
  3D APH (IoU@0.5) 
  \\ \hline
\multirow{6}{*}{\begin{tabular}[c]{@{}l@{}}LEVEL\_1/LEVEL\_2 \end{tabular}} &
  M3D-RPN\ddag~\cite{brazil2019m3d_rpn} &
  0.35/0.33 &
  0.34/0.33 &
  3.79/3.61 &
  3.63/3.46\\
  &
  PatchNet\dag \cite{patchnet} &
  0.39/0.38 &
  0.37/0.36 &
  2.92/2.42 &
  2.74/2.28\\
  &
  PCT \cite{pct} &
  0.89/0.66 &
  0.88/0.66 &
  4.20/4.03 &
  4.15/3.99\\
 &
  CaDDN \cite{reading2021categorical_caddn} &
  5.03/4.49 &
  4.99/4.45 &
  17.54/16.51&
  17.31/16.28\\
 &
  \emph{MonoFlex}$^{\ast}$ \cite{DBLP:conf/cvpr/ZhangL021_monoflex} &
  11.70/10.96 &
  11.64/10.90 &
  32.26/30.31 &
  32.06/30.12\\
 &
  \emph{DCD(Ours)} &
  \textbf{12.57}/\textbf{11.78} &
  \textbf{12.50}/\textbf{11.72} &
  \textbf{33.44}/\textbf{31.43} &
  \textbf{33.24}/\textbf{31.25}\\ \bottomrule
\end{tabular}}
\end{table}

\subsection{Ablation Studies}
\begin{figure}
\centering
\includegraphics[width=0.6\linewidth]{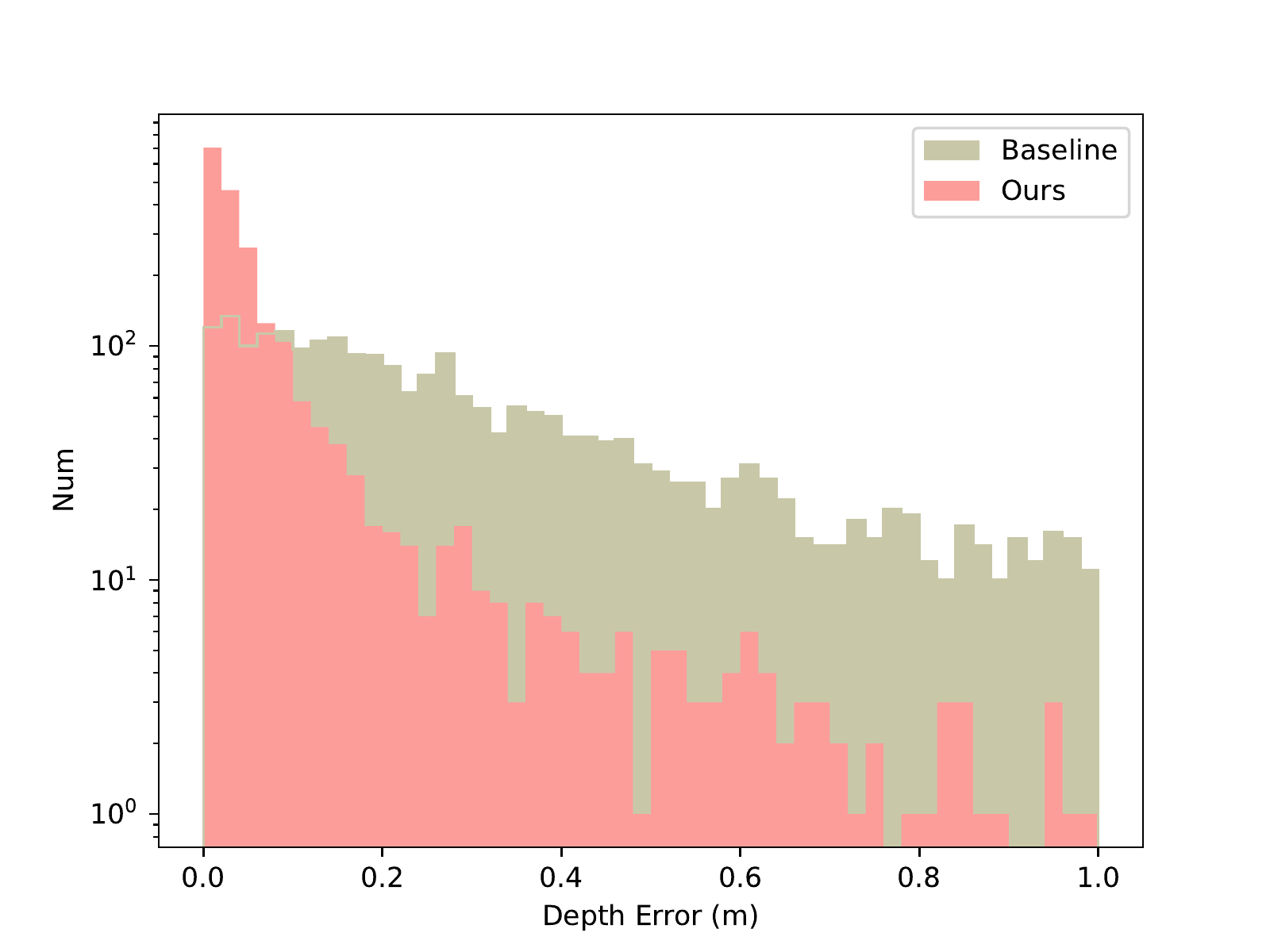}
\caption{This figure is the histogram of the depth error on the KITTI \emph{val} set. X-axis represents the distance from the predicted final depth to the GT depth, and Y-axis represents the number of depths of objects with log scale. As the figure shows, our method can estimate depth much more accurately than baseline.}
\label{fig:vis_depth_error}
% \end{wrapfigure}
\end{figure}

\noindent{\textbf{Keypoints enable better depth estimation.}}
We utilize multiple keypoints rather than bounding boxes to represent an object. The keypoints can accurately reflect the object's outline, which provides meaningful shape prior and abundant information for depth estimation. To show the benefits gained from dense geometric constraints, we visualize the predicted depths' error on KITTI \emph{val} set as Fig.~\ref{fig:vis_depth_error} shows.

% \begin{wrapfigure}{r}{.5\linewidth}

\noindent{\textbf{The effectiveness of DGDE.}} To discover the inner workings of multiple 2D-3D keypoints projection constraints, we apply the DGDE on the state-of-the-art method MonoFlex \cite{DBLP:conf/cvpr/ZhangL021_monoflex}. MonoFlex predicts eight bounding box corners and two top-down center points. In addition to the ten keypoints, we add another regression branch on MonoFlex to predict 63 semantic keypoints. Basically, they are sampled from the model surface representing the rough skeletons.

In Table \ref{tab:pairs and weighting}, the improvements from (d) to (e) are significant (+\textbf{1.30} on the Hard level) even with only 10 keypoints. With more keypoints ((c) and (f)), DGDE achieves holistic improvements on both the uncertainty-based weighting method and our matching-based method.

\noindent{\textbf{The more keypoints, the better performances.}}
Our depth estimator can produce depth candidates by edge projection constraints of arbitrary directions. To fully realize the potential of our depth estimator, we use all the extra 63 semantic keypoints. Thus, it is easy to generate numerous edges and obtain considerable depth candidates (2628).
With such a large number of depth candidates, it is more likely to generate an accurate and robust final depth. In Table \ref{tab:pairs and weighting}, the model (f) with all keypoints outperforms our baseline by a large margin of \textbf{2.31} AP on Easy level and \textbf{1.51} AP on Mod. level.

\noindent{\textbf{The effectiveness of \emph{Graph Matching Weighting module (GMW)}.}}
As the number of projection constraints increases, it is urgent to weigh the constraints appropriately since some are of low quality. To this end, we apply \emph{GMW} and compare it with uncertainty-based methods. The uncertainty-based methods ((a), (b) and (c)) estimate uncertainty independently for each edge. Thus they are not capable of exploiting the global instance information of all edge projection constraints. Different from that, \emph{GMW} is able to exploit the global information and achieve much better results. For example, model (f) (w/ GMW) surpasses model (c) (w/o GMW) by \textbf{1.55} AP on the Hard level, where the model is provided with global information to deal with severe occlusions.

\begin{table}[]
\caption{Quantitative results using the state-of-the-art method MonoFlex~\cite{DBLP:conf/cvpr/ZhangL021_monoflex} as baseline. This table shows the effectiveness of \emph{DGDE} and \emph{GMW}. The Sec.~\ref{sec:num_pairs} explains the strategy of choosing 1500 depth candidates.}
\label{tab:pairs and weighting}
\centering
\resizebox{\linewidth}{!}{
\begin{tabular}{l|l|c|c|c|ccc}
\toprule
\multirow{2}{*}{} & \multirow{2}{*}{Weighting   Method}& \multirow{2}{*}{\emph{DGDE}} & \multirow{2}{*}{\#Keypoints} & \multirow{2}{*}{\#Depth Candidates}  &  & $AP _{3D|R40|IoU@0.7} $ &  \\ \cline{6-8} 
    &                        &    &      &            & Easy  & Mod   & Hard  \\ \hline
(a) & Uncertainty (Baseline)~\cite{DBLP:conf/cvpr/ZhangL021_monoflex}
                             &           & 10 & 5     & 21.63 & 15.87 & 13.38 \\
(b) & Uncertainty            & \checkmark& 10 & 45    & 21.72 & 16.09 & 13.35 \\
(c) & Uncertainty            & \checkmark& 73 & 1500  & 22.84 & 16.53 & 13.77 \\
(d) & \emph{GMW}        &           & 10 & 5     & 22.58 & 16.14 & 13.63 \\
(e) & \emph{GMW}       & \checkmark& 10 & 45    & 23.30 & 16.91 & 14.93 \\
(f) & \emph{GMW}       & \checkmark& 73 & 1500  & \textbf{23.94} & \textbf{17.38} & \textbf{15.32} \\ 
\bottomrule
\end{tabular}}
\end{table}

\noindent{\textbf{Study of supervision priority in \emph{GMW}.}}
In Table~\ref{tab:reg_cls_loss}, we find that enabling depth regression supervision at the beginning detriments the performance. There is a straightforward explanation: when the match is incorrect, supervising the weighted depth will make the gradients noisy.

\begin{figure}
% \begin{wrapfigure}[20]{r}{.5\linewidth}
% \begin{figwindow}
\centering
\includegraphics[width=0.6\linewidth]{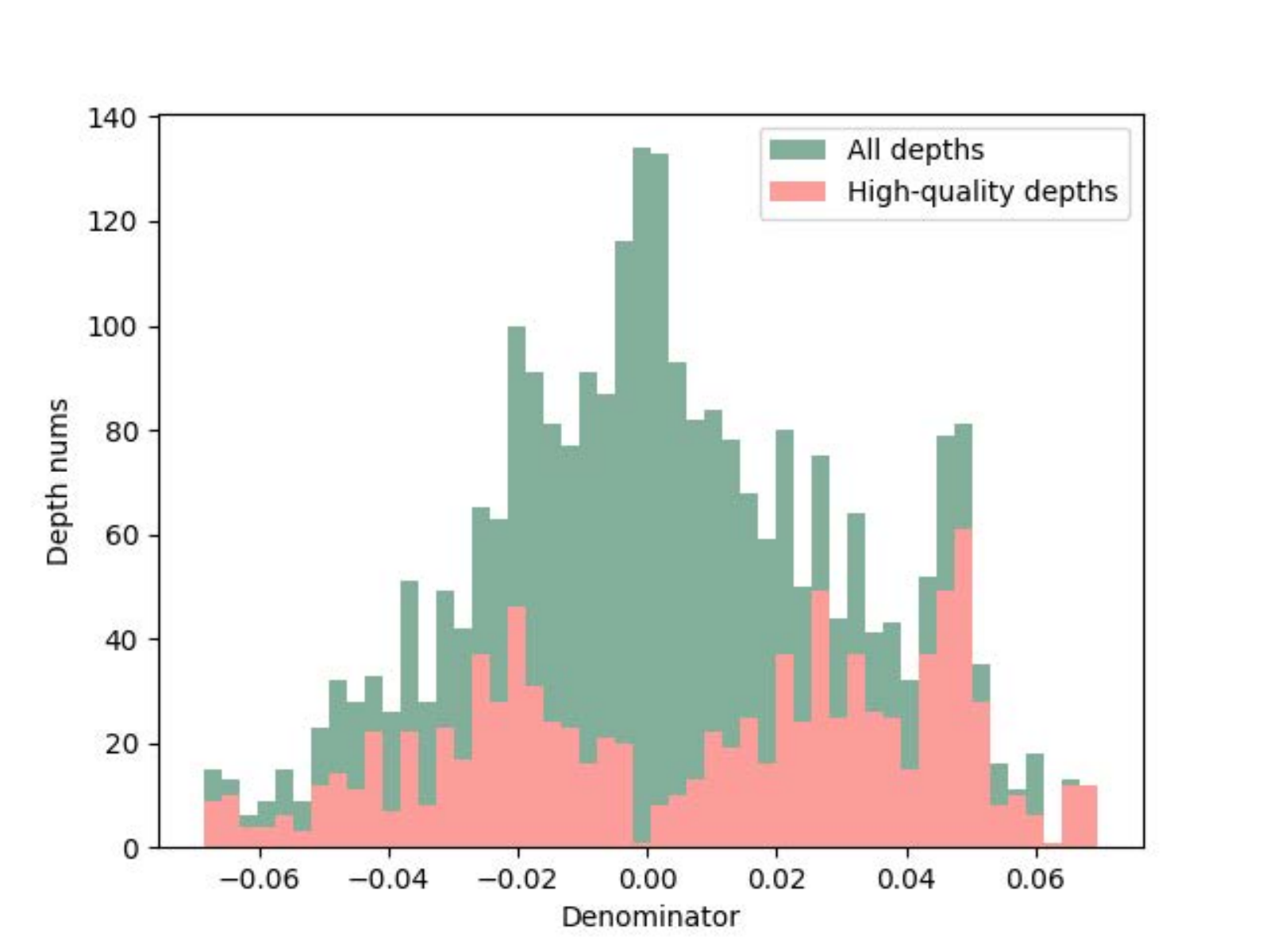}
\caption{The horizontal axis represents the value of the denominator of Eq.~\eqref{eq:zij} for an object. The heights of green bins represent the number of depth candidates computed by \emph{DGDE}, while the heights of red bins represent the number of high-quality depth candidates whose distance from ground truth depth is less than 0.5m.}
\label{fig:uv_mask}
\end{figure}

\noindent{\textbf{The Number of Edges.}}\label{sec:num_pairs}
The numerical calculation of the depth by  Eq.~\eqref{eq:zij} is very unstable when the denominator is too close to zero. We made a histogram for the analysis. As shown in Fig.~\ref{fig:uv_mask}, the vast majority of these small-denominator depths are of poor quality. For this reason, we use a mask to ignore the depth candidates with extremely small denominators. This ablation study shows how the number of selected depth candidates influences the model performance. As Table \ref{tab:number_edge} shows, the $AP\vert_{R40}$ increases when the number of selected depths is from 50 to 1500, peaks when the number of depths is 1500 and decreases afterward. 
% \end{wrapfigure}
% \end{window}
\begin{table}[]
\begin{center}
\noindent\begin{minipage}[t]{.6\linewidth}
\centering
\caption{Ablation of supervision priority of \emph{GMW} module.}
\label{tab:reg_cls_loss}
\resizebox{\linewidth}{!}{
\begin{tabular}{c|c|c|ccc}
\toprule
\multirow{2}{*}{Reg loss} & \multirow{2}{*}{Cls loss} & \multirow{2}{*}{Reg loss start} & \multicolumn{3}{c}{$AP_{3D|R40|IoU@0.7}$}                                                                    \\ \cline{4-6} 
                          &                           &                                       & Easy                            & Mod                             & Hard                            \\ \hline
                          & \checkmark & -                                     & 23.38                           & 17.03                           & 15.01                           \\
\checkmark & \checkmark & 0 epoch                                    & 22.93                           & 16.83                           & 14.72                           \\
\checkmark & \checkmark & 50 epochs                                    & \textbf{23.94} & \textbf{17.38} & \textbf{15.32} \\ \bottomrule
\end{tabular}}
\end{minipage}\hfill
\centering
\noindent\begin{minipage}[t]{.3\linewidth}
\caption{Ablation of the different number of edges.}
\label{tab:number_edge}
\centering
\resizebox{0.99\linewidth}{!}{
\begin{tabular}{c|ccc}
\toprule
\multirow{2}{*}{\#Edges} & \multicolumn{3}{c}{$AP_{3D|R40|IoU@0.7}$}                                                                  \\ \cline{2-4} 
                                     & Easy                            & Mod                             & Hard                            \\ \hline
50                                   & 23.35                           & 16.93                           & 15.01                           \\
500                                  & 23.58                           & 17.11                           & 15.13                           \\
1500                                 & \textbf{23.94} & \textbf{17.38} & \textbf{15.32} \\
2628                                 & 23.37                           & 16.98                           & 14.98                           \\ \bottomrule
\end{tabular}}
\end{minipage}
\end{center}
\end{table}

\subsection{Disscussion about DCD and AutoShape}
Although both of our method and AutoShape~\cite{liu2021autoshape} utilize multiple keypoints to estimate the object's location, there are three critical differences:
\begin{itemize}
    \item AutoShape directly uses all 2D-3D \emph{keypoints} projection constraints to solve the object depth. Our method solves a depth candidate from each \emph{edge} constraint. Thus, our edge constraints are not only in a larger number but also in higher order than the keypoint constraints.
    \item AutoShape generates keypoint weights independently without explicit interaction between keypoints. Our method uses a learnable graph matching module to model the edge constraints, so we produce each depth's weight based on all the edge constraints, leading to better weighting.
\end{itemize}
We re-implement Autoshape's depth estimation and weighting method on our baseline and the experiment result is in Table~\ref{tab:autoshape}.

\begin{table}[]
\centering
\caption{We re-implement the AutoShape's depth estimation and weighting regression methods on our baseline. The combined depth estimator combines all keypoint projection constraints as input and produces one depth as output. }
\label{tab:autoshape}
\resizebox{\linewidth}{!}{
\begin{tabular}{c|c|c|c|ccc}
\toprule
\multirow{2}{*}{Method}                                 & \multirow{2}{*}{\#Keypoints} & \multirow{2}{*}{Depth Estimatior} & \multirow{2}{*}{\#Depth Candidates} & \multicolumn{3}{c}{$AP_{3D|R40|IoU@0.7}$}                                                                    \\ \cline{5-7} 
                                                        &                           &                                   &                             & Easy                            & Mod                              & Hard                            \\ \hline
Autoshape \cite{liu2021autoshape} & 73                        & Combined                          & 1                           & 22.37                           & 16.48                            & 14.58                           \\
\emph{DCD} (Ours)                                                     & 73                        & \emph{DGDE}       & 1500                        & \textbf{23.94} & \textbf{ 17.38} & \textbf{15.32} \\ \bottomrule
\end{tabular}}
\end{table}

\section{Conclusion}
This paper proposes a method that can densely calculate an object's depth from 2D-3D projection constraints of edges of any direction. Therefore, we can obtain $n(n-1)/2$ depths for an object with $n$ keypoints. Moreover, we propose a novel framework that can generate reliable weights for each depth by matching the 2D-3D edges. We finally produce a robust depth by combing each depth candidate with its weight. The experiments show the effectiveness of our method, where we outperform all the existing methods in the KITTI and WOD benchmarks.

\section*{Acknowledgements}
\noindent{This work was supported in part by the Major Project for New Generation of AI (No.2018AAA0100400), the National Natural Science Foundation of China (No. 61836014, No. U21B2042, No. 62072457, No. 62006231). Also, our sincere and hearty appreciations go to Lue Fan, who polishes our paper and offers many valuable suggestions.}

% \clearpage
% \unboldmath
\newpage
\bibliographystyle{splncs04}
\bibliography{egbib}

\newpage
\appendix
{\noindent \textbf{\LARGE{Appendix}}}
\section{WOD detailed results}
We also provide the detailed results on WOD~\cite{waymo} as Table~\ref{tab:my-table} and Table~\ref{tab:my-table2} show.
\begin{table}[]
\centering
\caption{The IoU@0.7 result on WOD~\cite{waymo} \emph{val} set. \emph{Italics}: These methods utilize the whole \emph{train} set, while the others uses $1/3$ amount of images in \emph{train} set. \ddag: M3D-RPN is re-implemented by \cite{reading2021categorical_caddn}. \dag: PatchNet is re-implemented by \cite{pct}. $^{\ast}$: MonoFlex is our baseline and re-implemented ourselves.}
\resizebox{0.8\linewidth}{!}{
\begin{tabular}{l|c|cccc|cccc}
\toprule
\multirow{2}{*}{Difficulty} &
  \multirow{2}{*}{Method} &
  \multicolumn{4}{c|}{3D mAP} &
  \multicolumn{4}{c}{3D mAPH} \\
 &
   &
  Overall &
  0-30m &
  30-50m &
  50-$\infty$ &
  Overall &
  0-30m &
  30-50m &
  50m-$\infty$ \\ \hline
\multirow{4}{*}{\begin{tabular}[c]{@{}l@{}}LEVEL\_1\\ @IoU 0.7\end{tabular}} &
  M3D-RPN\ddag~\cite{brazil2019m3d_rpn} &
  0.35 &
  1.12 &
  0.18 &
  0.02 &
  0.34 &
  1.10 &
  0.18 &
  0.02 \\
  &
  PatchNet\dag \cite{patchnet} &
  0.39 &
  1.67 &
  0.13 &
  0.03 &
  0.37 &
  1.63 &
  0.12 &
  0.03 \\
  &
  PCT \cite{pct} &
  0.89 &
  3.18 &
  0.27 &
  0.07 &
  0.88 &
  3.15 &
  0.27 &
  0.07 \\
 &
  CaDDN \cite{reading2021categorical_caddn} &
  5.03 &
  14.54 &
  1.47 &
  0.10 &
  4.99 &
  14.43 &
  1.45 &
  0.10 \\
 &
  \emph{MonoFlex$^{\ast}$} \cite{DBLP:conf/cvpr/ZhangL021_monoflex} &
  11.70 &
  30.64 &
  5.29 &
  1.05 &
  11.64 &
  30.48 &
  5.27 &
  1.04 \\
 &
  \emph{DCD(Ours)} &
  \textbf{12.57} &
  \textbf{32.47} &
  \textbf{5.94} &
  \textbf{1.24} &
  \textbf{12.50} &
  \textbf{32.30} &
  \textbf{5.91} &
  \textbf{1.23} \\ \hline
\multirow{4}{*}{\begin{tabular}[c]{@{}l@{}}LEVEL\_2\\ @IoU 0.7\end{tabular}} &
  M3D-RPN\ddag~\cite{brazil2019m3d_rpn} &
  0.33 &
  1.12 &
  0.18 &
  0.02 &
  0.33 &
  1.10 &
  0.17 &
  0.02 \\
  &PatchNet\dag \cite{patchnet}&
  0.38&
  1.67&
  0.13&
  0.03&
  0.36&
  1.63&
  0.11&
  0.03\\
  &PCT \cite{pct}&
  0.66&
  3.18&
  0.27&
  0.07&
  0.66&
  3.15&
  0.26&
  0.07\\
 &
  CaDDN \cite{reading2021categorical_caddn} &
  4.49 &
  14.50 &
  1.42 &
  0.09 &
  4.45 &
  14.38 &
  1.41 &
  0.09 \\
 &
  \emph{MonoFlex$^{\ast}$} \cite{DBLP:conf/cvpr/ZhangL021_monoflex} &
  10.96 &
  30.54 &
  5.14 &
  0.91 &
  10.90 &
  30.37 &
  5.11 &
  0.91 \\
 &
  \emph{DCD(Ours)} &
  \textbf{11.78} &
  \textbf{32.30} &
  \textbf{5.76} &
  \textbf{1.08} &
  \textbf{11.72} &
  \textbf{32.19} &
  \textbf{5.73} &
  \textbf{1.08} \\ \bottomrule
\end{tabular}}
\label{tab:my-table}
\end{table}

\begin{table}[]
\centering
\caption{The IoU@0.5 result on WOD\cite{waymo} \emph{val} set. \emph{Italics}: These methods utilize the whole \emph{train} set, while the others uses $1/3$ amount of images in \emph{train} set. \ddag: M3D-RPN is re-implemented by \cite{reading2021categorical_caddn}. \dag: PatchNet is re-implemented by \cite{pct}. $^{\ast}$: MonoFlex is our baseline and re-implemented ourselves.}
\resizebox{0.8\linewidth}{!}{
\begin{tabular}{l|c|cccc|cccc}
\toprule
\multirow{2}{*}{Difficulty} &
  \multirow{2}{*}{Method} &
  \multicolumn{4}{c|}{3D mAP} &
  \multicolumn{4}{c}{3D mAPH} \\
 &
   &
  Overall &
  0-30m &
  30-50m &
  50-$\infty$ &
  Overall &
  0-30m &
  30-50m &
  50m-$\infty$ \\ \hline
\multirow{4}{*}{\begin{tabular}[c]{@{}l@{}}LEVEL\_1\\ @IoU 0.5\end{tabular}} &
  M3D-RPN\ddag~\cite{brazil2019m3d_rpn} &
  3.79 &
  11.14 &
  2.16 &
  0.26 &
  3.63 &
  10.70 &
  2.09 &
  0.21 \\
  &
  PatchNet\dag \cite{patchnet} &
  2.92 &
  10.03 &
  1.09 &
  0.23 &
  2.74 &
  9.75 &
  0.96 &
  0.18 \\
  &
  PCT \cite{pct} &
  4.20 &
  14.70 &
  1.78 &
  0.39 &
  4.15 &
  14.54 &
  1.75 &
  0.39 \\
 &
  CaDDN \cite{reading2021categorical_caddn} &
  17.54 &
  45.00 &
  9.24 &
  0.64 &
  17.31 &
  44.46 &
  9.11 &
  0.62 \\
 &
  \emph{MonoFlex$^{\ast}$} \cite{DBLP:conf/cvpr/ZhangL021_monoflex} &
  32.26 &
  61.13 &
  25.85 &
  9.03 &
  32.06 &
  60.75 &
  25.71 &
  8.95 \\
 &
  \emph{DCD(Ours)} &
  \textbf{33.44} &
  \textbf{62.70} &
  \textbf{26.35} &
  \textbf{10.16} &
  \textbf{33.24} &
  \textbf{62.35} &
  \textbf{26.21} &
  \textbf{10.09} \\ \hline
\multirow{4}{*}{\begin{tabular}[c]{@{}l@{}}LEVEL\_2\\ @IoU 0.5\end{tabular}} &
  M3D-RPN\ddag ~\cite{brazil2019m3d_rpn}&
  3.61 &
  11.12 &
  2.12 &
  0.24 &
  3.46 &
  10.67 &
  2.04 &
  0.20 \\
  &PatchNet\dag  \cite{patchnet}&
  2.42&
  10.01&
  1.07&
  0.22&
  2.28&
  9.73&
  0.94&
  0.16\\
  &PCT \cite{pct}&
  4.03&
  14.67&
  1.74&
  0.36&
  3.99&
  14.51&
  1.71&
  0.35\\
 &
  CaDDN \cite{reading2021categorical_caddn} &
  16.51 &
  44.87 &
  8.99 &
  0.58 &
  16.28 &
  44.33 &
  8.86 &
  0.55 \\
 &
  \emph{MonoFlex$^{\ast}$} \cite{DBLP:conf/cvpr/ZhangL021_monoflex} &
  30.31 &
  60.91 &
  25.11 &
  7.92 &
  30.12 &
  60.54 &
  24.97 &
  7.85 \\
 &
  \emph{DCD(Ours)} &
  \textbf{31.43} &
  \textbf{62.48} &
  \textbf{25.60} &
  \textbf{8.92} &
  \textbf{31.25} &
  \textbf{62.13} &
  \textbf{25.46} &
  \textbf{8.86} 
  \\ \bottomrule
\end{tabular}}
\label{tab:my-table2}
\end{table}

\end{document}